\documentclass{article} 
\usepackage{iclr2016_conference,times}
\usepackage{hyperref}
\usepackage{url}
\usepackage[pdftex]{graphicx}
\usepackage{natbib}
\usepackage{textcomp}

\title{GradNets: Dynamic Interpolation Between Neural Architectures}

\author{Diogo Almeida\thanks{Equal contribution. Author ordering determined by coin flip.}, Nate Sauder\footnotemark[1] \\
Enlitic\\
\texttt{\{diogo,nate\}@enlitic.com} \\
}

\begin{document}

\maketitle

\begin{abstract}
In machine learning, there is a fundamental trade-off between ease of optimization and expressive power. Neural Networks, in particular, have enormous expressive power and yet are notoriously challenging to train. The nature of that optimization challenge changes over the course of learning. Traditionally in deep learning, one makes a static trade-off between the needs of early and late optimization. In this paper, we investigate a novel framework, GradNets, for dynamically adapting architectures during training to get the benefits of both. For example, we can gradually transition from linear to non-linear networks, deterministic to stochastic computation, shallow to deep architectures, or even simple downsampling to fully differentiable attention mechanisms. Benefits include increased accuracy, easier convergence with more complex architectures, solutions to test-time execution of batch normalization, and the ability to train networks of up to 200 layers.
\end{abstract}

\section{Introduction}
Deep learning has become the state of the art system for many machine learning challenges (\cite{krizhevsky2012imagenet}, \cite{girshick2014rich}). Many of deep learning\textquotesingle s benefits come from its ability to automatically learn abstract representations from raw input features, its scalability to large datasets, and its exponential growth in expressive power with depth. As a consequence of this great expressive power, deep learning can suffer from slow convergence, over-fitting, poor optima, difficulty of optimization, and acute sensitivity to initialization and hyper-parameters. Deep learning practitioners often optimize for the best possible generalization.

Dropout and rectified linear units (ReLUs) are two of the more important algorithmic breakthroughs of the last 5 years. Indeed, both were essential to the breakthrough results of \citet{krizhevsky2012imagenet} that reduced the error rate on the ImageNet Challenge of 2012 by almost 40\%. Both of these techniques sacrifice ease of optimization early in training in exchange for expressive power. Dropout\textquotesingle s addition of stochastic noise improves performance by enabling larger networks and preventing co-adaptation of weights albeit at a significant cost of convergence speed \citep{hinton2012improving}. Despite being easier to optimize than sigmoids or hyperbolic tangents, ReLUs \citep{nair2010rectified} increase the expressive power of networks while being more difficult to initialize and optimize than linear models \citep{saxe2013exact} or leaky ReLUs \citep{maas2013rectifier}.

Traditionally, architectures are chosen before training and remain static throughout training. Machine learning practitioners often choose expressive power and generalizability over ease of training in these static architectures.

GradNets, dynamic interpolation between architecture decisions during training, allows for architectures that maintain the desired expressive power and generalizability but also benefit from easier optimization.

In particular, applying the GradNets framework to both dropout and ReLUs have led to improved accuracy on CIFAR-10, convergence with levels of dropout that static networks diverge on, and successful training of multi-layer perceptrons of up to 200 layers on MNIST through smoother information flow.

\section{Method Description}
\begin{figure}
  \centering
    \includegraphics[scale=0.5]{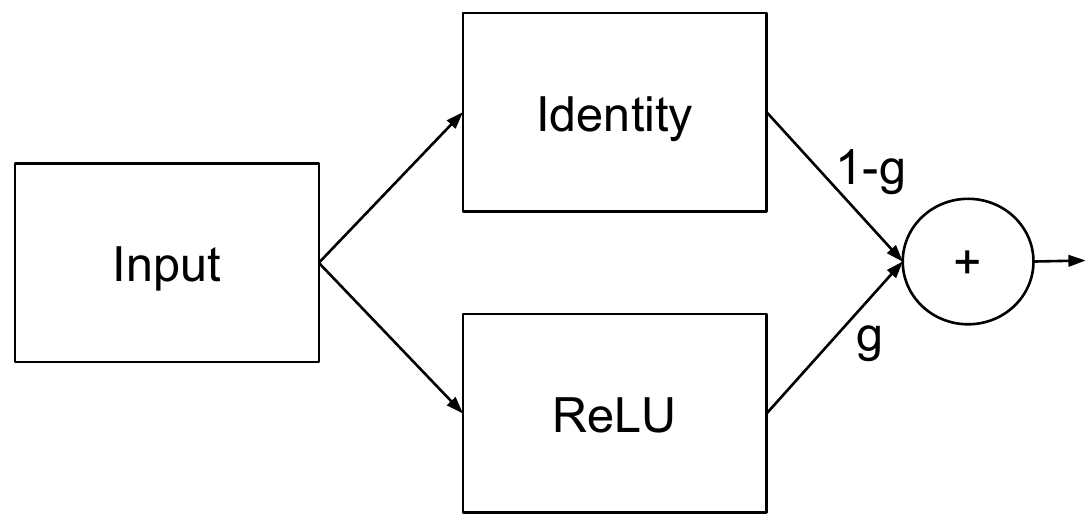}
    \caption{Architectural diagram of GReLU.}
  \label{fig:gradnet-diagram}
\end{figure}

A GradNet architectural component is exceptionally simple: a weighted mean of two architectural components where the weight $g$ is annealed from 0 to 1 over training. For example, a gradual ReLU anneals from an identity layer to a ReLU layer (see Figure \ref{fig:gradnet-diagram}).

In this paper, we use a linear schedule with a single hyperparameter $\tau$ as the epoch when the annealing is complete:

$$ g = min(t / \tau, 1) $$

where $t$ is the epoch number.

The GradNet framework is exceptionally flexible as it can be applied to any pair of network components that result in a tensor of the same shape. Our examples can be found in Table~\ref{table:gradnet-techniques}.

\begin{table}[t]
\begin{center}
\begin{tabular}{lll}
\multicolumn{1}{c}{\bf Technique}  &\multicolumn{1}{c}{\bf Early Component} &\multicolumn{1}{c}{\bf Late Component}
\\ \hline
Gradual ReLU (GReLU) & Identity & ReLU \\
Inverse GReLU & Absolute Value & ReLU \\
Gradual Dropout & Identity & Dropout \\
Gradual Mean-Max Pooling & Mean Pool & Max Pool \\
Gradual Batch Normalization & Batch Normalization & Identity \\
Gradual Convolution & Identity & Convolution \\
Gradual NiN & Convolution & Convolution + 1x1 Convolution \\
Gradual Spatial Transformer Network & Mean Pool & Spatial Transformer Network \\
\end{tabular}
\caption{Different applications of GradNet framework, along with corresponding early and late components.}
\label{table:gradnet-techniques}
\end{center}
\end{table}

\section{Experiments}

\subsection{Baseline Architecture}

\subsubsection{CIFAR-10}
Experiments in sections \ref{initial-cifar} through \ref{final-cifar} were performed on CIFAR-10. Data augmentation was applied with a 50\% probability to horizontally flip the input image. We employ the network topology of “Scalable Bayesian Optimization Using Deep Neural Networks” \citep{snoek2015scalable}. When not using gradual dropout, activations after each convolutional layer were dropped out with probability 0.3. When Batch Normalization was used, the mean and variance at test time were attained by taking an exponential moving average of train time means and variances. The network was optimized using Adam with a batch size of 256 and early stopping and initialized using orthogonal initialization. $\tau$ for GradNets was set to be 100 epochs.

\subsubsection{MNIST}
Experiments in section \ref{only-mnist} were performed on MNIST. A multilayer perceptron was used with 512 units per hidden layer, which was optimized with Adam with a batch size of 500 and early stopping and initialized using orthogonal initialization. $\tau$ for GradNets was set to be 5 epochs.

\subsubsection{Cluttered MNIST}

Experiments in section \ref{only-cluttered-mnist} were performed on Cluttered MNIST\citep{mnih2014recurrent}. An architecture with 2 convolutional layers, each 3x3 with 32 filters followed by a max pooling layer, and 2 fully connected layers, with 256 and 10 units respectively, was used to classify 20x20 pixel images. The input 60x60 pixel images were downsampled either with 3x3 mean pooling or an affine Spatial Transformer Network \citep{jaderberg2015spatial}, whose localization network had an architecture of 2 convolutional layers, each 5x5 with 20 filters followed by a max pooling layer, and 2 fully connected layers, with 50 and 10 units respectively with 512 units per hidden layer, Models were optimized with Adam with a batch size of 500 and early stopping and initialized using orthogonal initialization. $\tau$ for GradNets was set to be 5 epochs.

\subsection{Gradual ReLUs (GReLUs)}
\label{initial-cifar}

ReLUs are the dominant form of nonlinearity in modern deep neural networks, however they do not allow gradients to flow well leading to several proposed solutions: leaky and very leaky ReLUs or ReLUs with parameterized slope for the negative part \citep{he2015delving}. On the other hand, linear networks are easier to optimize due to their convexity, but don\textquotesingle t allow for the learning of abstract hierarchical feature detectors. By beginning with fully linear networks and interpolating towards fully ReLU networks, we gain the ease of optimization of linear networks and also the expressive power of ReLU networks. We call them Gradual ReLUs.

Instead of interpolating between two separate outputs, it would be equivalent to interpolate the slope of the negative part of leaky ReLUs resulting in a technique with no overhead over traditional ReLU networks.

Table~\ref{table:grelu-cifar10} shows that applying GReLUs to a large CNN architecture shows that they outperform not only ReLUs, but also their leaky, very leaky, and parameteric variants.

In addition to testing the interpolation between a network with no nonlinearity and a ReLU, we also interpolated between an absolute value and ReLU (see Inverse GReLU in Table~\ref{table:grelu-cifar10}). This is equivalent to annealing from a ReLU leak of -1 instead of 1. The standard GReLU outperforms this architecture, showing that the initial linearity is important, but this architecture also outperforms standard ReLUs. One possible reason for this is that having a network be dynamic changes the trajectory of optimization for the better.


\begin{table}[t]
\begin{center}
\begin{tabular}{ll}
\multicolumn{1}{c}{\bf Nonlinearity}  &\multicolumn{1}{c}{\bf Valid Accuracy}
\\ \hline
ReLU	& 0.8686 \\
Leaky ReLU	& 0.8615 \\
Very Leaky ReLU	& 0.6347 \\
PReLU & 0.8728 \\
\hline
GReLU	& \textbf{0.8823} \\
Inverse GReLU	& 0.8767 \\
\end{tabular}
\caption{GReLU Accuracy on CIFAR-10}
\label{table:grelu-cifar10}
\end{center}
\end{table}

\subsection{Gradual Dropout}
Stochastic dropout is an excellent regularizer but negatively affects convergence speed so much so that some state of the art architectures even remove it \citep{ioffe2015batch}. Furthermore, with sufficiently high levels of dropout, networks don\textquotesingle t train at all. Thus, there is a balance between regularizing networks that are easily overfit and the ability for a network to be optimized. In gradual dropout, beginning training with zero dropout and annealing up to higher levels (even as high as 0.9) means that early training from randomly initialized weights does not suffer from excessive noise while the final networks can be sufficiently large yet regularized to generalize well to other domains. 

\begin{figure}
  \centering
    \includegraphics[scale=0.5]{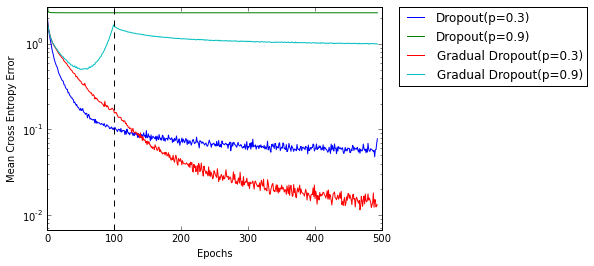}
    \caption{Training error with and without Gradual Dropout. The dashed vertical line indicates the epoch at which $g = 1$. Figure demonstrates that gradual dropout identifies weights that can continue to be optimized even when dropout is increased to very high levels  resulting in better local optima.}
  \label{fig:gradual-dropout-cifar10}
\end{figure}

Table~\ref{table:gradual-dropout-cifar10} shows that, similarly to with GReLUs, applying gradual dropout to the same CNN architecture shows that it not only outperforms using a static amount of dropout, but allows using much larger amounts of dropout to gain better performance. This is despite $\tau$ appearing to be too low - the results for $p=0.9$ show that the GradNet interpolates too aggressively.

A future extension of this technique could be used to find the optimal amount of dropout for a network in a single training run by seeing at which point the network\textquotesingle s performance starts to drop, instead of relying on an expensive hyperparameter search.

\begin{table}[t]
\begin{center}
\begin{tabular}{ll}
\multicolumn{1}{c}{\bf Model}  &\multicolumn{1}{c}{\bf Valid Accuracy}
\\ \hline
Dropout(p=0.3)	& 0.8686 \\
Gradual Dropout(p=0.3)	& 0.8732 \\
Dropout(p=0.5)	& 0.8756 \\
Gradual Dropout(p=0.5)	& \textbf{0.8793} \\
Dropout(p=0.7)	& 0.8173 \\
Gradual Dropout(p=0.7)	& 0.8613 \\
Dropout(p=0.9)	& 0.0973 \\
Gradual Dropout(p=0.9)	& 0.8311 \\
\end{tabular}
\caption{Gradual Dropout Accuracy on CIFAR-10}
\label{table:gradual-dropout-cifar10}
\end{center}
\end{table}

\subsection{Gradual Pooling}

Max pooling is the most commonly used subsampling approach for state of the art networks. However, low level features are not informative early in training and thus max pooling obstructs gradient flow to most of these early units. To address this problem, we use the same effect of annealing from linear to non-linear networks. In particular, we capture this intuition by annealing from mean pooling units to max pooling units. As seen in Table~\ref{table:gradual-pooling-cifar10}, this combined approach does better than either mean or max pooling as well as taking an average, learning a weighted average, or learning an attention mechanism \citep{lee2015generalizing}.

\begin{table}[t]
\begin{center}
\begin{tabular}{ll}
\multicolumn{1}{c}{\bf Model}  &\multicolumn{1}{c}{\bf Valid Accuracy}
\\ \hline
Max Pool (Baseline)	& 0.8686 \\
Mean Pool & 0.8517 \\
Constant Mixed Pooling	& 0.8696 \\
Gated Pooling	& 0.8699 \\
Learnable Mixed Pooling	& 0.8705 \\
Gradual Mean to Max	& \textbf{0.8743} \\
\end{tabular}
\caption{Gradual Pooling on CIFAR-10}
\label{table:gradual-pooling-cifar10}
\end{center}
\end{table}

\subsection{Gradual Batch Normalization}
Batch Normalization (BN) is a state of the art architectural component that greatly aids convergence and accuracy. However, it cannot easily be executed at test time, especially when evaluating on single data points. There are several clever ways of computing means and variances to work around this fundamental problem however, ideally, we would like the convergence and accuracy benefits of BN while avoiding issues at validation time. Using the GradNet framework to anneal from BN layers to identity layers, we generate a network that is trivial to evaluate at validation time while exploiting all the benefits of batch normalization.

\begin{table}[t]
\begin{center}
\begin{tabular}{ll}
\multicolumn{1}{c}{\bf Model}  &\multicolumn{1}{c}{\bf Valid Accuracy}
\\ \hline
w/o Batch Normalization	& 0.8686 \\
w/ Batch Normalization	& 0.8739 \\
Gradual Batch Normalization	& \textbf{0.8904} \\
\end{tabular}
\caption{Gradual Batch Normalization on CIFAR-10}
\label{table:gradual-bn-cifar10}
\end{center}
\end{table}

\subsection{Gradual Convolutions}
Shallower networks are easier to train early but less powerful than deep networks. Accordingly, it is intuitively appealing to leverage the easier convergence of shallower networks while maintaining the power of very deep networks. Two proposed solutions, Highway networks \citep{srivastava2015highway} and Network in Network (NiN) \citep{lin2013network} are emblematic of this problem: NiN is very deep but challenging to train while Highway networks require much larger depths than normal to match baseline performance resulting in greatly increased computational cost. For the depth used, highway networks perform significantly worse than the baseline network while interpolating from identity layers into convolutional layers (thereby progressively adding depth) results in almost no loss of performance. On the other hand, interpolating from convolutional layers into NiN layers results in significant improvements in performance. 

\begin{table}[t]
\begin{center}
\begin{tabular}{ll}
\multicolumn{1}{c}{\bf Model}  &\multicolumn{1}{c}{\bf Valid Accuracy}
\\ \hline
Conv (baseline)	& 0.8686 \\
Highway Conv	& 0.7635 \\
Gradual Conv	& 0.8618 \\
Gradual NiN	& \textbf{0.8787} \\
\end{tabular}
\caption{Gradual Convolutions on CIFAR-10}
\label{table:gradual-conv-cifar10}
\end{center}
\end{table}

\subsection{Combining GradNets}
\label{final-cifar}
In Table~\ref{table:combining-gradnets-cifar10}, we demonstrate the composability of GradNets.

\begin{table}[t]
\begin{center}
\begin{tabular}{ll}
\multicolumn{1}{c}{\bf Model}  &\multicolumn{1}{c}{\bf Valid Accuracy}
\\ \hline
Baseline	& 0.8686 \\
GReLU	& 0.8811 \\
Gradual Dropout p=0.5	& 0.8777 \\
GReLU + Gradual Dropout p=0.5	& \textbf{0.8884} \\
\end{tabular}
\caption{Combining GradNets on CIFAR-10}
\label{table:combining-gradnets-cifar10}
\end{center}
\end{table}

\subsection{GReLU Depth Experiments}
\label{only-mnist}

The trend in neural networks has been towards ever deeper architectures, which is aligned with the theoretical results indicating that they have exponentially greater expressibility. However, deeper networks exacerbate the challenges we have highlighted. For example, poor information flow in ReLU networks. As shown on Figure \ref{fig:gradual-dropout-cifar10}, using the GradNet framework not only improves the performance for slightly deeper networks but also allows for the successful training of significantly deeper networks which standard networks do not successfully train.

\begin{figure}
  \includegraphics[width=0.5\linewidth]{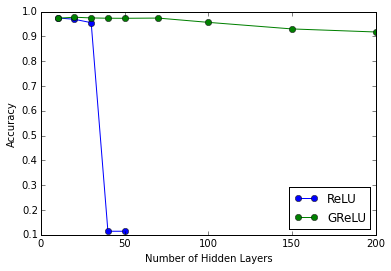}
  \includegraphics[width=0.5\linewidth]{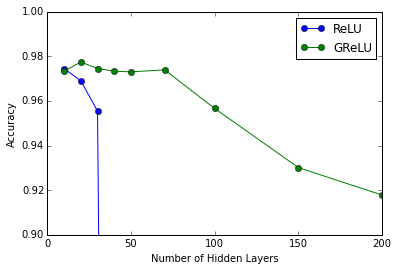}
  \caption{Validation accuracy on MNIST as the number of hidden layers in the MLP increases.}
  \label{fig:grelu-deep-mnist}
\end{figure}

\subsection{Gradual Spatial Transformer Networks}
\label{only-cluttered-mnist}

Spatial transformer networks \citep{jaderberg2015spatial} are a powerful new type of attentional model that makes performing an affine glimpse differentiable. This makes it significantly easier to train than reinforcement learning methods however it still struggles with convergence. In particular, when randomly initialized, the localization network can transform the original image off the screen or zoom in excessively. 

Since the localization network depends on the signal from the classification network, it makes sense to train the classification network first before tuning the localization network. The flexibility of the GradNet framework allows for a computational implementation of this intuition. In particular, we gradually anneal from a simple downsample to an affine spatial transformer network.

As shown in Table~\ref{table:gradual-stn-cluttered-mnist}, the regular spatial transformer often diverges. However the GradNet implementation does not diverge and yet accomplishes the same top level results.

\begin{table}[t]
\begin{center}
\begin{tabular}{lll}
\multicolumn{1}{c}{\bf Model}  &\multicolumn{1}{c}{\bf Avg. Valid Accuracy} &\multicolumn{1}{c}{\bf \% Diverging}
\\ \hline
Mean Pool	& 0.874 & \textbf{0} \\
Affine STN	& 0.5375 (0.962) & 50 \\
GradNet Mean Pool to Affine STN	& \textbf{0.961} & \textbf{0} \\
\end{tabular}
\caption{Gradual Spatial Transformer Networks on Cluttered MNIST. Parenthesized result for Affine STN is mean accuracy for models that converge. \% divergence was determined over 10 runs.}
\label{table:gradual-stn-cluttered-mnist}
\end{center}
\end{table}

\section{Related Work}

\subsection{Preconditioning}
\citet{hinton2006reducing} achieved breakthrough results by training deeper multi-layer networks through preconditioning early layers by greedy training each layer in succession. More recently, \citet{yosinski2014transferable} initialized from known successful weights thereby dropping the network immediately into a lower point of the loss surface and avoiding getting stuck in the many local optima that arise with deeper and larger networks. Distillation \citep{hinton2015distilling} is another technique that uses existing cumbersome models to extract structure from the data before transferring it to a less cumbersome model.

GradNets embody this philosophy by uniting stages of training that are easy to optimize and powerful with the additional benefit that training can be done jointly in a single training phase.

\subsection{Rectified Linear Units (ReLU) Alternatives}
Plenty of ReLU alternatives have been proposed and indeed some have been integral to winning competitions. For the most part, these experiments have shown that networks are robust to the slope for the negative part. Whereas ReLUs saturate all gradients on the negative side to zero; leaky ReLUs allow that information to flow. Accordingly, very leaky ReLUs have been extremely successful at deep learning competitions. PReLUs \citep{he2015delving} instead allow for training the slope on the negative edge of the ReLU thus allowing the network to train how linear its nonlinearities should be. In randomized ReLUs \citep{xu2015empirical}, slopes are randomly chosen within a specific range before being frozen during testing. The ensuing stochastic regularization proved successful on the Kaggle data science bowl competition.

Rather than making a static trade-off between ease of optimization and expressive power, GradNets allow us to dynamically change this trade-off during training.

\subsection{Very Deep Networks}
Highway networks \citep{srivastava2015highway} are compositions of layers where a gating tensor controls the interpolation between an identity layer and a regular fully connected layer. The smooth varying enables learning much deeper networks however the computation of the gating factor at each layer requires twice the computation compared to standard networks.

\citet{saxe2013exact} showed the benefits of using deep linear networks as a medium for studying the learning dynamics of deep neural networks. In particular, they found that with precise initializations they could achieve depth independent learning times for deep linear networks.

GradNets leverage insights from both these results: borrowing the theoretically justified initializations for linear networks while gaining the optimization benefits of the interpolation operation in highway networks without incurring their additional computational cost. These results speak to GradNet\textquotesingle s ability to exploit the theoretical insights of simpler architectures while training very complex architectures.

\subsection{Attentional Models}
Interpolation between architectures is very similar to the soft attention used to interpolate between differentiable data structures (\cite{joulin2015inferring}, \citet{grefenstette2015learning}). These use backpropagation to learn the weights for interpolation.

However, soft attention incurs the computational cost of computing all possible outputs instead of controlling interpolation to converge to a single output.

\section{Conclusion}
A difficult choice in neural network design is navigating the fundamental trade-off between ease of optimization and expressiveness. GradNets are a framework to help this choice by exploiting the ease of optimization early in training and the expressiveness late in training. Our experiments show that not only does this allow greater amounts of stochastic regularization as well as successful training of extremely deep networks but it also consistently improves performance at almost no computational cost. Finally, the results on spatial transformer networks speak to the ability of GradNets to dramatically ease training of complex architectural components and to enable new classes of components to be used in neural networks.

Future work includes interpolating between any simpler, easier to train architectural components and a more complex expressive component that generalizes better. Due to the results on deeper networks, GradNets may provide enormous benefits to recurrent architectures. Furthermore, the framework itself can be extended by adapting the interpolation policy using smarter techniques, such as reinforcement learning.

\subsubsection*{Acknowledgments}
We thank NVIDIA for their generosity in providing access to part of their cluster in support of Enlitic\textquotesingle s mission and our research. 

\bibliography{iclr2016_conference}
\bibliographystyle{iclr2016_conference}

\end{document}